\documentclass{new_tlp}
\usepackage{mathptmx}
\usepackage{multirow}
\newcommand\bcmdtab{\noindent\bgroup\tabcolsep=0pt%
  \begin{tabular}{@{}p{10pc}@{}p{20pc}@{}}}
\newcommand\ecmdtab{\end{tabular}\egroup}

  \title[Theory and Practice of Logic Programming]
        {Improving Adherence to Heart Failure Management Guidelines via Abductive 
        	Reasoning}

  \author[Z. Chen, E. Salazar, K. Marple, L. Tamil, G. Gupta, S. Das, A. Amin]
         {Zhuo Chen, Elmer Salazar, Kyle Marple, Gopal Gupta, Lakshman Tamil\\
         University of Texas at Dallas, Texas, USA.\\
         \email{$\{$zxc130130,ees101020,kbm072000,gupta,laxman$\}$@utdallas.edu}
         \and Sandeep Das, Alpesh Amin\\
         Cardiology Division, Department of Internal Medicine, University of 
         Texas Southwestern Medical Center, Texas, USA.\\
         \email{$\{$Sandeep.Das,Alpesh.Amin$\}$@UTSouthwestern.edu}
        }

\jdate{March 2003}
\pubyear{2003}
\pagerange{\pageref{firstpage}--\pageref{lastpage}}
\doi{S1471068401001193}

\begin{document}
\nocite{*}% includes all entries of BibTeX database into the list of references.

\label{firstpage}

\maketitle

  \begin{abstract}
   Management of chronic diseases such as heart failure (HF) is a major public 
   health problem. A standard approach to managing chronic diseases by medical 
   community is to have a committee of experts develop guidelines that all 
   physicians should follow. Due to their complexity, these guidelines are 
   difficult to implement and are adopted slowly by the medical community at large. We have developed a physician advisory system that codes the entire set of 
   clinical practice guidelines for managing HF using answer set programming 
   (ASP). In this paper we show how abductive reasoning can be deployed to find 
   missing symptoms and conditions that the patient must exhibit in order for a 
   treatment prescribed by a physician to work effectively. Thus, if a physician 
   does not make an appropriate recommendation or makes a non-adherent 
   recommendation, our system will advise the physician about symptoms and 
   conditions that must be in effect for that recommendation to apply. It is under consideration for acceptance in TPLP.
  \end{abstract}


%\tableofcontents

\section{Introduction}
Heart failure is the inability of the heart to keep up with the demands of the 
body. As a result, an inadequate amount of blood is supplied to the human body 
and/or pressures in the heart can rise. This can cause congestion of blood in 
the lungs, abdomen, legs. All of this culminates in symptoms of exercise 
intolerance. Half of the people diagnosed with HF die within five years. In 
U.S. alone there are 5.7 million people currently living with HF 
\cite{Heart_disease_and_stroke_statistics}.

Optimal management of HF requires adherence to evidence-based clinical practice 
guidelines. The American College of Cardiology Foundation (ACCF)/ American Heart 
Association (AHA) Guidelines for the Management of Heart Failure have been 
created by a multi-disciplinary committee of experts and are based on thorough 
review of best available clinical evidence on the management of heart failure. 
They represent a consensus among experts on the appropriate treatment and 
management of HF \cite {Clinical_Practice_Guideline_Methodology_Summit_Report}.

Although evidenced-based guidelines should be the basis for all disease 
management, physicians' adherence to those guidelines is poor 
\cite{Why_don't_physicians_follow}. One of the reasons for this is that 
guidelines can be quite complex, as is the case for HF management. For example, 
more than 100 variables have been associated with mortality and 
re-hospitalization in heart failure patients. In the 2013 ACCF/AHA Guideline 
for the Management of Heart Failure \cite{guideline}, the variables range from 
straightforward information like age and sex to sophisticated data like 
electrocardiogram results and the history of HF-related symptoms and diseases. 
There are more than 60 complex rules to integrate all these data together. Such 
complexity can lead to delays in adoption and missed opportunities at giving 
effective recommendations for even the most experienced healthcare providers 
\cite{Enhancing_the_use_of_clinical_guidelines}.

We have developed a physician advisory system that assists physicians in 
adhering to the guidelines for managing HF \cite{iclp2016} based on ASP 
\cite{Truszczynski,Ilkka}.  This system 
automates the 2013 ACCF/AHA Guideline for the Management of Heart Failure and 
gives recommendations similar to human physicians that strictly follows the 
guidelines, even if the information about the patient is incomplete. 
This system has been implemented using the the s(ASP) system
\cite{sasp}, a goal-directed ASP system 
\cite{Goal-directed_execution_of_answer_set_programs}. Moreover, knowledge patterns in the guideline were identified and modularized using ASP. An example of the application of knowledge patterns in the system can be found in Sect. \ref{sec:PAS}. 
The outputs the physician advisory system produces have been found to be consistent with our interpretation of the guidelines. The system has also been tested with the help of our cardiologist co-authors on representative patient data provided by them. 
These tests showed that the system makes treatment recommendations 
consistent with physicians' recommendations. In fact, in some cases the system 
made treatment recommendations that resulted in physicians revising their 
original treatment recommendations.

In this paper we build on our previous work and show how abductive reasoning can be 
employed in our system to help physicians validate their treatment recommendations. 
In practice, physicians often need to give treatment plans based on incomplete 
information about a patient, especially in underdeveloped areas where complete 
information about a patient is not always available. We show how our physician advisory 
system helps to ensure the compliance with the 2013 ACCF/AHA Guideline for the 
Management of Heart Failure via abductive reasoning. {\it The core idea is that the 
(non-adherent) physician-prescribed treatment is treated as an observation and fed as a query 
to our physician advisory system, which is then run in the abductive mode to 
discover the necessary evidence that must be available for the given treatment 
to be applicable}. This evidence may be symptoms that the patient must exhibit
or the conditions that must apply to the patient for the (non-adherent) physician-prescribed treatment to be applicable. It is likely that the 
physician overlooked establishing this evidence, which is now produced via abduction. In light of this abduced evidence, the physician may want to re-evaluate his/her treatment recommendation(s). 

Case studies based on this methodology and its results are 
also presented. Given that compliance with the guidelines is 
poor \cite{Why_don't_physicians_follow}, we show how 
abduction can be used to inform a physician about symptoms and conditions that 
must exist in a patient for a non-guideline-compliant treatment recommendation 
to be applicable. Our abduction-powered system can also be used as a tool for training physicians for managing HF.  
The system has been evaluated on ten representative patients, data for which
was provided by our cardiologist co-authors. In four of these ten cases our system pointed out symptoms/conditions that may have been overlooked. In three of the ten cases, our system could not abduce anything to validate the original recommendation, indicating that the recommendation should be rejected or closely examined. In three of the cases, the physician's recommendation were correct, i.e., compliant with the guidelines.

Our research makes the following contribution: 

\begin{itemize}
\item We show how abductive reasoning within ASP can be useful 
for medical diagnosis, especially, for heart failure management.
To the best of our knowledge, our work is the first real-life application of abduction with ASP technology in the very important area of medicine.

\item Our research illustrates a major advantage of goal-directed ASP systems such as s(ASP), namely, their ability to perform effective abductive reasoning. As explained later, it is difficult to determine the minimum set of abducibles with SAT-solver based ASP systems such as CLASP. In \cite{iclp2016}, we briefly mentioned s(ASP)'s potential to produce abducibles with respect to a query. This paper demonstrates the technical details of the implementation of abductive reasoning in s(ASP). Ten verified experimental results using abduction are also presented.

\end{itemize}

\section{Abductive Reasoning In Goal-directed ASP Systems} \label{sec:Abductive 
	Reasoning In Goal-directed ASP Systems}
Our physician advisory system has been developed on the s(ASP) Predicate Answer 
Set Programming system. The s(ASP) system also performs abduction under the 
stable model semantics. We give a brief introduction to both goal-directed 
execution of answer set programs and abductive reasoning for answer set 
programming. We assume that the reader is familiar with ASP.

%\subsection{Goal-directed ASP Systems} \label{subsec:Goal-directed ASP Systems}

\smallskip
\noindent{\bf Goal-directed Execution of ASP:}
With the ability to compute stable models of normal logic programs, 
goal-directed systems for executing answer set programs adopt a different 
approach than traditional ASP systems that are almost all based on the use of 
SAT solvers. Traditional ASP systems find the entire model (answer set) of a 
given answer set program. Checking if a query Q is satisfied in this program 
then amounts to checking if Q is present in the computed answer set. In 
contrast, a goal-directed system is query driven. Given an answer set program P 
and a query goal Q, 
a goal-directed system for executing answer set programs will systematically 
enumerate all answer sets that contain the propositions/predicates in Q.
This enumeration employs Coinductive SLD resolution (or Co-SLD resolution), 
which systematically computes elements of the greatest fixed point (GFP) of a 
program via backtracking \cite{Goal-directed_execution_of_answer_set_programs}. 
With s(ASP), our goal-directed method lifts the restriction that answer set 
programs must be finitely groundable 
\cite{Goal-directed_execution_of_answer_set_programs}.

One ramification of the top-down execution strategy of a goal-directed ASP 
system is that the answer set produced may be partial 
\cite{Goal-directed_execution_of_answer_set_programs}.\footnote{Because answer
	sets are partial, we enumerate elements that are true as well as false in
	the partial answer set.} Consider the following 
program:

\begin{verbatim}
p :- not q.          r :- not s.
q :- not p.          s :- not r.
\end{verbatim}

Using a goal-directed ASP system, the query :- q will produce $\left\{q,\ not\ 
p\right\}$ as the answer because the rules containing r and s are independent 
of the rules for q and p. Had the query been :- q, s, the system would give 
$\left\{q,\ not\ p,\ s,\ not\ r\right\}$ as the result. Hence the part of the 
answer set we get is determined by the query. The partial answer set essentially
contains the elements that are necessary to establish the query.

The Galliwasp system \cite{Galliwasp} was the first efficient implementation of 
the goal-directed method. It is essentially an implementation of A-Prolog \cite{BARAL} 
that uses grounded normal logic programs as input. The underlying algorithm of 
Galliwasp makes use of coinduction \cite{Coinductive_Logic_Programming} to find 
partial answer sets containing a given query. In query-driven, top-down execution, coinduction is helpful in processing {\it even loops}  \footnote{An even loop occurs when a recursive call is encountered with an even, non-zero number of negations between the call and its ancestor in the call stack. \cite{sasp_paper}} where cyclical nature of execution needs to be detected and used as a basis for goal-success \cite{Galliwasp}.
Specifically, Galliwasp uses call 
graphs to classify rules based on two attributes: (i) a rule is said to contain 
an \emph{odd loop over negation} (OLON) if it can be called recursively with an 
odd number of negations between the initial call and its recursive invocation, 
and (ii) it is called an ordinary rule if it has at least one path in the call 
graph that will not result in such a call. Galliwasp uses 
ordinary rules to generate 
candidate answer sets and OLON rules to reject invalid candidate sets. An 
implementation of Galliwasp is available \cite{Galliwasp}.

The s(ASP) predicate ASP system \cite{sasp_paper,sasp} operates in a manner similar to its 
predecessor, Galliwasp, in that rules are classified according to negation (OLON 
rules and non-OLON rules) and executed in a goal-directed manner. 
However, s(ASP) completely removes the need 
to ground input programs. Therefore, unlike Galliwasp and other ASP solvers, 
s(ASP) programs do not need to have a finite grounding. That is, the s(ASP) 
system executes \textit{predicate} answer set programs \textit{directly, 
	without grounding them at all}. Unlike traditional ASP systems, s(ASP) programs 
can contain arbitrary terms; lists, terms and complex data structures can 
appear as arguments of predicates in s(ASP) programs. Thus, s(ASP) is a much 
more powerful and expressive system. The introduction of full predicates with terms requires a number of innovative techniques to support them. These techniques include co-SLD resolution \cite{Coinductive_Logic_Programming}, constructive negation, and dual rules \cite{sasp_paper,sasp}. An implementation of the 
s(ASP) system is available \cite{sasp} at sourceforge. 

\medskip

\noindent{\bf Abduction}:
The term \emph{abduction} refers to a form of reasoning that is concerned 
with the generation and evaluation of explanatory hypotheses. Abductive 
reasoning leads back from facts to a proposed explanation of those facts.
According to \cite{harman_abduction}, abductive reasoning takes the following 
form:

\begin{verbatim}
      The fact B is observed.
      But if A were true, B would be a matter of course.
      Hence, there is reason to suspect that A is true.
\end{verbatim}

In this form, \texttt{B} can be either a particular event or an empirical 
generalization. \texttt{A} serves an explanatory hypothesis and \texttt{B} 
follows from \texttt{A} combined with relevant background knowledge. Note that 
\texttt{A} is not necessarily true, but plausible and worthy of further 
validation. A simple example of abductive reasoning is that one might attribute 
the symptoms of a common cold to a viral infection.

\subsection{Abductive Reasoning in ASP}
More formally, abduction is a form of reasoning where, given the premise $P 
\Rightarrow Q$, and the observation $Q$, one surmises
({\it abduces}) that $P$ holds \cite{kakas}. More generally, given a theory 
$T$, an observation
$O$, and a set of abducibles $A$, then $E$ is an explanation of $O$ (where $E 
\subset A$) if:
\smallskip

\begin{enumerate}
	\item $T\cup E \models O$
	\item $T\cup E $ is consistent
\end{enumerate}

%\smallskip % KM note: removed; seems useless
\noindent Generally, $A$ consists of a set of propositions such that
if ${\tt p} \in A$, then there is no rule in $T$ with {\tt p} as its head. We
assume the theory $T$ to be an answer set program. Under a goal-directed 
execution
regime, an ASP system can be extended with abduction by simply adding the 
following
rule:

\begin{verbatim}
p :- not _not_p.
_not_p :- not p.
\end{verbatim}

\noindent for every ${\tt p} \in A$.
Posing the observation $O$ as a query to this extended answer set program
will yield all the explanations as part of a (partial) answer set. The reason 
why
this works is simple: rules of the form above have two answer sets, one in 
which {\tt p} is false 
and another in which {\tt p} is true.
These settings (along with assignments to other propositions) may make the
observation $O$ succeed. As an example, consider the program below:

\begin{verbatim}
p :- a, not q.
q :- a, b.
q :- c.
\end{verbatim}

\noindent and the abducibles {\tt $\{$a, b, c$\}$}. If we add the clauses below 
to program above:

\begin{verbatim}
a :- not _not_a.        b :- not _not_b.
_not_a :- not a.        _not_b :- not b.
c :- not _not_c.
_not_c :- not c.
\end{verbatim}

\noindent then the query (observation) {\tt ?- p} will succeed, producing the
answer set {\tt $\{$p, not q, a, not b, not c$\}$} and abducibles
{\tt $\{$a, not b, not c$\}$}. First, {\tt \_not\_a, \_not\_b} and {\tt 
	\_not\_c} are not included in the 
answer set as they are auxiliary predicates introduced for convenience. Note that s(ASP) omits from output the predicates starting with an underscore ('\_').
Second, in goal-directed execution, 
since the models
generated may be partial,  propositions/predicates that are known to be true, 
as well as
propositions/predicates known to be false, are shown explicitly in the 
(partial) answer set. 

It should also be noted that abductive reasoning is more precise under 
goal-directed execution. Other execution
strategies, most of which are based around SAT-solvers (e.g., the CLASP system 
\cite{Clingo}), compute the entire stable model, which can produce confusing 
answers where it is not clear as to which of the abducibles are involved in the 
explanation. Consider the example program above. Suppose we extend the set of 
abducibles to include a new proposition {\tt $\{$d$\}$} and add the two 
customary rules involving {\tt d} and {\tt not\_d}. For observation {\tt 
	$\{$p$\}$}, if we use a traditional ASP solver, now we will get two answer 
sets: {\tt $\{$p, not q, a, not b, not c, d$\}$} and {\tt $\{$p, not q, a, not 
	b, not c, not d$\}$}. 
These two answer sets essentially are extensions of the single answer set 
produced by the goal-directed method,
however, this answer set has to be replicated twice---once for including {\tt 
	$\{$d$\}$} and once for including {\tt $\{$not d$\}$}. Thus, if there are 
irrelevant abducibles, then for a given observation, the number of answer sets 
will proliferate 
exponentially. Considerable analysis will be required to extract the true 
abducibles. The goal-directed execution method does not suffer from this 
problem, as it only explores the space of knowledge that is relevant to the 
observation. Thus, only relevant propositions/predicates are abduced.

\section{Physician Advisory System for HF management} \label{sec:PAS}
In this section we discuss the guidelines for HF management, as well as give an 
overview of our physician advisory system \cite{iclp2016}.

The 2013 ACCF/AHA Guideline for the Management of Heart Failure is 
based on four progressive stages of heart failure. Stage A includes patients at 
risk of heart failure who are asymptomatic and do not have structural heart 
disease. Stage B describes asymptomatic patients with structural heart 
diseases; it includes New York Heart Association (NYHA) class I, in which 
ordinary physical activity does not cause symptoms of heart failure. Stage C 
describes patients with structural heart disease who have prior or current 
symptoms of heart failure; it includes NYHA class I, II (slight limitation of 
physical activity), III (marked limitation of physical activity) and IV (unable 
to carry on any physical activity without symptoms of heart failure, or 
symptoms of heart failure
at rest). Stage D describes patients with refractory heart failure who require 
specialized interventions; it includes NYHA class IV. Interventions at each 
stage are aimed at reducing risk factors (stage A), treating structural heart 
disease (stage B) and reducing morbidity and mortality (stages C and D).

The treatment recommendations in these four stages are articulated in the ACCF/AHA Class of Recommendation (COR) and Level of Evidence (LOE). COR is an estimate of the size of the treatment effect considering risks versus benefits in addition to evidence and/or agreement that a given treatment or procedure is or is not useful/effective or in some situations may cause harm. COR is summarized as follows:

\begin{itemize}
	\item Class I recommendation: the treatment is effective and should be performed.
	
	\item Class IIa recommendation: the treatment can be useful and is probably recommended.
	
	\item Class IIb recommendation: the treatment's usefulness is uncertain and it might be considered.
	
	\item Class III recommendation: the treatment is harmful or unhelpful and should be not administrated.
\end{itemize}

One rule for treatment recommendations in the guideline looks as follows:

\begin{quote}
	{\noindent ``Aldosterone receptor antagonists are recommended to reduce 
		morbidity and mortality following an acute MI in patients who have LVEF of 40\% 
		or less who develop symptoms of HF or who have a history of diabetes mellitus, 
		unless contraindicated.''}
\end{quote}

To overcome the difficulties 
that physicians face in implementing the guidelines, we have developed a 
physician advisory system that automates the 2013 ACCF/AHA Guideline for the 
Management of Heart Failure. Our physician advisory system is able to give 
recommendations with justifications like a real human physician who strictly 
follows the guidelines, even under the condition of incomplete information 
about the patient. 

The physician advisory system for HF management consists of the rule 
database and a fact table. 
%CLASP system \cite{Clingo} and s(ASP) system. 
The rule database covers all the knowledge in the 2013 ACCF/AHA Guideline for the 
Management of Heart Failure. The fact table contains the relevant information 
of the patient with heart failure such as demographics, history of present 
illness, daily symptoms, medication intolerance, known risk factors, 
measurements as well as ACCF/AHA stage and NYHA class \cite{guideline}. These
rules and facts can be loaded into the s(ASP) system, and the query {\tt ?- 
	recommendation(Treatment, Recommendation\_class)} posed to find possible treatments for this patient. There may be more than one potential treatment that can be discovered
via backtracking. 
%The s(ASP) system has a justification feature,
%which allows a user to examine the justification (essentially a proof trace)
%for each recommendation.

The rules above can be grounded, and then the grounded program can be run 
on the CLASP system \cite{Clingo} as well. Note that SAT-based ASP systems such
as CLASP will report all the recommendations \textit{in a single answer set}, as they
cannot distinguish between treatments that are in the same model but are
independent of each other. However, knowing all applicable recommendations
for a patient is useful, as will be discussed later. These applicable recommendations 
are pharmaceutical treatments, management objectives and device/surgical therapies. s(ASP) is able to generate all valid recommendations for a given 
patient as well, however, CLASP is much faster in terms of execution
efficiency.
%while s(ASP) shines in preserving the information 
%during the reasoning process. 
The main advantage of the s(ASP) system for our HF application is that it is able to generate
individual recommendations separately along with all the predicates/propositions
needed to support that recommendation. 
%It is also able to generate abducibles as reminders to physicians by performing 
%abductive reasoning in cases where a recommendation is non-adherent 
%to guidelines.

The guideline rules are fairly complex and require the use of negation as 
failure, non-monotonic reasoning and reasoning with incomplete information. A 
fairly common situation in medicine is that a drug can only be recommended if 
its use is not contraindicated (i.e., the use of the drug will not have an 
adverse impact on that patient). Contraindication is naturally modeled via 
negation as failure. The ability of answer set programming to model defaults, 
exceptions, weak exceptions, preferences, etc., makes it ideally suited for 
coding these guidelines. Not surprisingly, the program for the physician advisory system is unstratified. Both odd and even loops over negation occur in the program. As an example, the indispensable choice knowledge pattern \cite{iclp2016} used in our system contains an even loop. As a concrete example of non-stratification, consider the following rule
from ACCF/AHA stage C \cite{guideline}:

\begin{quote}
	{\noindent ``In patients with a current or recent history of fluid 
		retention, beta blockers should not be prescribed without diuretics.''}
\end{quote}

\noindent The pattern describes a case in which if a choice is made, some other choices must also be made; if those choices can't be made, then the first choice is revoked \cite{iclp2016}. Note that Prolog conventions are followed (variables begin with an upper case letter).

\begin{verbatim}
  recommendation(beta_blockers, class_1) :-
     not absent_indispensable_choice(beta_blockers),
     not rejection(beta_blockers), evidence(accf_stage_c),
     diagnosis(hf_with_reduced_ef).
	
  absent_indispensable_choice(beta_blockers) :-
     not recommendation(diuretics, class_1), 
     diagnosis(hf_with_reduced_ef), evidence(accf_stage_c),       
     current_or_recent_history_of_fluid_retention.
	
  recommendation(diuretics, class_1) :-
   	 recommendation(beta_blockers, class_1),  
   	 diagnosis(hf_with_reduced_ef),
   	 not rejection(diuretics), evidence(accf_stage_c),
   	 current_or_recent_history_of_fluid_retention.
\end{verbatim}

This paper focuses on the physician advisory system's ability to perform 
abductive reasoning. Using this abductive capability, a 
physician can, for example, figure out the symptoms that 
a particular patient must exhibit for a given treatment recommended by
the doctor (so that this treatment is consistent 
with the 2013 ACCF/AHA Guideline for the Management of Heart Failure). The 
automatic checking for compliance using the physician advisory system is 
discussed in detail in Sect. \ref{sec:automatic checking for compliance}.

\section{Automatic Checking of Adherence to Guidelines} 
\label{sec:automatic checking for compliance}
\noindent{\bf Abductive Reasoning in s(ASP)}: Abductive reasoning is supported 
by the s(ASP) system, where the abducibles 
predicates are specified as follows, for example:

{\tt \#abducible goal(X).}

\noindent This declaration means that the specified goal may be abduced when 
failure of the query would otherwise occur during execution. Specifically, 
s(ASP) creates the following rules for the declaration statement above:

\begin{verbatim}
goal(X) :- not _neggoal(X), abducible(goal(X)).
_neggoal(X) :- not goal(X).
abducible(goal(X)) :- not _negabducible(goal(X)).
_negabducible(goal(X)) :- not abducible(goal(X)).
\end{verbatim}

\noindent{\bf Case Study}:
As mentioned earlier, many times available patient information is incomplete or 
guidelines are not complied with. A physician may give prescriptions using 
his/her intuitions based on incomplete information about a patient. In such a
case, the physician's recommendation can be run in the s(ASP) system in the
abductive mode where the physician's recommendation is posed as  a query. The system will then tell the physician, via abduction, any symptoms, conditions,
or required evidence that must be present for the recommendation to 
be correct. The physician can 
then work on ascertaining the presence of those 
symptoms, conditions, etc., or revise their
recommendation if those symptoms/conditions are not present. So in this case, while the physician may have made an incorrect recommendation, it can still be applicable if the abduced symptoms/conditions/evidence were to be present.  In contrast, if the query representing the physician's recommendation fails, then it means that the physician's recommendation is incorrect with respect to the guidelines, and should be rejected or carefully re-evaluated. 

Our aim, thus, is 
to help improve compliance with the clinical practice guidelines via abductive 
reasoning. 
%Our physician advisory system can be used to check if a 
%treatment 
%recommendation is compliant with the clinical practice guidelines. 
This is 
achieved as follows: all guideline-compliant treatment recommendations are 
first generated in a single answer set with the help of the CLASP system. Any 
treatment recommendation made by a doctor that is not among 
the recommendations made by CLASP system is considered a 
non-guideline-compliant treatment recommendation. A non-guideline-compliant 
treatment recommendation made by a physician is posed as an observation and fed 
as a query to s(ASP) system, which is then run in the abductive mode to 
discover the necessary evidence that must be available for the given treatment 
to be applicable.

Figure \ref{fig:patient information} displays the fact table distilled from the profile of 
one of the 10 representative patients. The representative patients' data were 
provided by the University of Texas Southwestern Medical Center. 

\begin{figure}
	\figrule
	\begin{center}
\begin{verbatim}
%doctor's assessments                %demographics
evidence(accf_stage_c).              evidence(female).
evidence(nyha_class_3).              evidence(age, 60).
%history of the patient
diagnosis(hf_with_reduced_ef).       diagnosis(diabetes).
diagnosis(dilated_cardiomyopathy).   evidence(angina).
history(standard_neurohumoral_antagonist_therapy).
%measurements from the lab                     
measurement(potassium, 3.0).         measurement(lvef, 0.16).
measurement(creatinine, 1.49).       measurement(non_lbbb, 110).
measurement(nt_pro_bnp, 5051).
measurement(glomerular_filtration_rate, 44).  
%contraindications
contraindication(crt).     contraindication(ace_inhibitors).

\end{verbatim}
\end{center}
\caption{Representation of the patient's information}\label{fig:patient information}
\figrule
\end{figure}
%\vspace{-0.3cm}
%\centerline{{\bf Figure 1:} Representation of the patient's information}

\smallskip
According the patient's profile, her doctor prescribed the combination of hydralazine and isosorbide dinitrate (isordil/hydralazine) to this patient, However, we know that the 
recommendation of isordil/hydralazine is not reasonable for this patient under
the guidelines once we run the CLASP system on the data. 

Since isordil/hydralazine is highly effective in treating HF for 
African Americans, we want to make sure we did not miss any vital information 
regarding their recommendation. The applicable rules for this class I recommendation of 
isordil/hydralazine is reproduced below \cite{guideline}:

\begin{quote}
	{\noindent ``Class I : The combination of hydralazine and isosorbide 
		dinitrate is recommended to reduce morbidity and mortality for patients 
		self-described as African Americans with NYHA class III-IV HFrEF receiving 
		optimal therapy with ACE inhibitors and beta blockers , unless 
		contraindicated.''}
\end{quote}

\begin{quote}
	{\noindent ``The combination of hydralazine and isosorbide dinitrate 
		should not be used for the treatment of HFrEF in patients who have no prior use 
		of standard neurohumoral antagonist therapy and should not be substituted for 
		ACE inhibitor or ARB therapy in patients who are tolerating therapy without 
		difficulty.''}
\end{quote}

\noindent Now we declare all possible abducibles which are not shown in the patient's profiles (this step can be automated using simple string matching):

\begin{verbatim}
#abducible contraindication(beta_blockers).
#abducible contraindication(icd).
...
#abducible diagnosis(atrial_fibrillation).
#abducible diagnosis(hypertension).
...
#abducible evidence(fluid_retention).
#abducible evidence(sleep_apnea).
...
#abducible history(angioedema).
#abducible history(thromboembolism).
...
\end{verbatim}

\noindent Abducibles involving numeric values are declared as textual propositions, e.g., ``lvef\_less\_than\_30'', ``mi\_post\_40\_days'', etc. And corresponding auxiliary rules are introduced. For example:

\begin{verbatim}
lvef_less_than_30:- measurement(lvef, Data), Data =< 30.
mi_post_40_days:- measurement(mi, Data), Data >= 40. 
\end{verbatim}

\noindent The declaration of abducibles should be placed after facts and rules in s(ASP) since we do not want to abduce known facts or deducible facts. This arrangement guarantees that whatever s(ASP) abduces leads us to overlooked or missed evidence which is indispensable to justify the treatment recommendation in question.

Then we pose the following query (the observation) to the s(ASP) system
with the abductive reasoning flag set:

\begin{verbatim}
recommendation(hydralazine_and_isosorbide_dinitrate, class_1)
\end{verbatim}

With the abducibles declared, s(ASP) may abduce them when failure of the query 
would otherwise occur. In this case, s(ASP) returns false, which means there is 
truly nothing we can do to make class I recommendation of isordil/hydralazine,  
given the patient's data.

Next we want to know if it is possible to make class IIa recommendation of 
isordil/hydralazine. The relevant rules in the guideline \cite{guideline} 
regarding the class IIa recommendation of isordil/hydralazine are shown below:

\begin{quote}
	{\noindent ``Class IIa : A combination of hydralazine and isosorbide 
		dinitrate can be useful to reduce morbidity or mortality in patients with 
		current or prior symptomatic HFrEF who cannot be given an ACE inhibitor or ARB 
		because of drug intolerance, hypotension, or renal insufficiency, unless 
		contraindicated.''}
\end{quote}

\noindent Next, we pose the following query (the observation) to s(ASP) running
in abductive mode:

\begin{verbatim}
recommendation(hydralazine_and_isosorbide_dinitrate, class_2a)
\end{verbatim}

\noindent One of the answer sets as well as the list of abducibles given by s(ASP) are displayed in Fig. \ref{fig:Results of abductive reasoning}. 

\begin{figure}
\figrule
\begin{center}
\begin{verbatim}

{ contraindication(ace_inhibitors), contraindication(arbs),
 diagnosis(hf_with_reduced_ef), evidence(accf_stage_c), history(ace_inhibitors),
 recommendation(hydralazine_and_isosorbide_dinitrate,class_2a),
 treatment(hydralazine_and_isosorbide_dinitrate), 
 not absent_indispensable_treatment(hydralazine_and_isosorbide_dinitrate), 
 not contraindication(hydralazine_and_isosorbide_dinitrate), 
 not rejection(hydralazine_and_isosorbide_dinitrate), 
 not skip_concomitant_treatment(hydralazine_and_isosorbide_dinitrate), 
 not superseded(hydralazine_and_isosorbide_dinitrate), 
 not taboo_choice(hydralazine_and_isosorbide_dinitrate) }
 
Abducibles: { history(angioedema), contraindication(arbs) }
\end{verbatim}
\end{center}
\caption{Results of abductive reasoning} \label{fig:Results of abductive reasoning}
\figrule
\end{figure}
%\centerline{{\bf Figure 2:} Result of abductive reasoning}

We immediately know from Fig. 2 that two things are 
abduced by the system: 1) the contraindication of ARBs and 2) the history of 
angioedema. It is clear from the Fig. 1 that both of the 
abducibles are not in the patient's profile. With some medical knowledge, we 
know that the history of angioedema is an indication for the contraindication 
of ACE inhibitors, which is listed in the fact sheet. It is reasonable for us 
to make a class IIa recommendation once we confirm the intolerance of ARBs for 
this patient. In this case, the patient's profile says nothing about her 
intolerance of ARBs. Thus, the doctor could look for this piece of information 
elsewhere to ensure that she does not miss useful information regarding the 
recommendation of isordil/hydralazine. Once she has done that, she will have 
the justification for the original recommendation of isordil/hydralazine.

Our experiments were done on a set of ten representative patients. 
For these ten patients, we had detailed patient data, as well as physician's
recommendation.
Table \ref{table: 10 cases} displays the experimental results for seven of 
these patients. The remaining three patients are not shown in the table 
because the physician's prescriptions (recommendation) for them are 
compliant with the guidelines. Thus there is no need 
to perform abductive reasoning for them. The first column presents the patient 
number. The second column shows the non-guideline-compliant treatment 
recommendations given by physicians. The third column displays the 
abducibles (symptoms, conditions, etc.) that must be established before
the recommendation shown in the second column can be regarded as compliant 
with the guidelines. It is worth noting that for patients No.4, No.8 and No.10, the system produced a failure (i.e., it failed to abduce anything).  This implies that the original recommendation made by the physician in each of these cases was incorrect. In the case No.4, the patient data shows no sign of atrial fibrillation or cardioembolic source, which makes anticoagulants unhelpful. In both case No.8 and No.10, the latest lab results say their potassium level is greater than 5.0 mEq/L, which makes aldosterone antagonists harmful to them. The cardiologist co-authors of this paper agree with the system's conclusions for these cases.

%Since known facts contributed to the rejection of what was recommended by the %physicians, the physician must re-evaluate the case carefully. 

\begin{table}
	\centering
	\caption{Results of abductive reasoning for 10 representative cases} 
	\label{table: 10 cases}
	\begin{tabular}{lll} 
		
		\hline
		\hline
		\textbf{Patient Number} & \textbf{Non-guideline-compliant}  & 
		\textbf{Abducibles} \\
		& \textbf{recommendations} & \\
		\hline
		\multirow{2}{*}{ No.1} 
		& \multirow{2}{*}{isordil/hydralazine (class IIa)} & 
		contraindication(arbs). \\
		& & history(angioedema). \\
		\hline
		\multirow{1}{*}{ No.4} 
		& \multirow{1}{*}{anticoagulants (class I)} & 
		None found \\
		
		\hline
		\multirow{1}{*}{ No.5} 
		& \multirow{1}{*}{implantable cardioverter defibrillator  (class I)} & survival\_year\_greater\_than\_1.\\
		
		\hline
		\multirow{2}{*}{ No.6} 
		& \multirow{2}{*}{isordil/hydralazine (class IIa)} & 
		contraindication(arbs). \\
		& & history(angioedema). \\
		\hline
		\multirow{1}{*}{ No.7} 
		& \multirow{1}{*}{aldosterone antagonists (class I)} & 
		evidence(recent\_mi). \\
	
		\hline
		\multirow{1}{*}{ No.8} 
		& \multirow{1}{*}{aldosterone antagonists (class I)} & 
		None found \\
		\hline
		\multirow{1}{*}{ No.10} 
		& \multirow{1}{*}{aldosterone antagonists (class I)} & 
		None found \\
		\hline
		\hline

	\end{tabular}
	\smallskip
	
\end{table}

Table \ref{table: running time} presents the running time of the physician advisory system to obtain all applicable treatment recommendations using CLASP and performing abductive reasoning via s(ASP) for all ten cases. As can be seen, the execution times are quite acceptable.

\begin{table}

	\centering
	\caption{Running time of the physician advisory system} 
	\label{table: running time}
	\begin{tabular}{lll} 
		
		\hline
		\hline
		\textbf{Patient Number} & \textbf{Obtaining applicable recommendations}  & 
		\textbf{Computing abducibles } \\
		& \textbf{(CLASP)} & \textbf{(s(ASP))} \\ 
		\hline
		No.1 & 0.021s & 3.451s \\
		\hline
		No.2 & 0.021s & N/A  \\
		\hline
		No.3 & 0.020s & N/A \\
		\hline
		No.4 & 0.020s & 0.965s \\
		\hline
		No.5 & 0.020s & 3.440s \\
		\hline
		No.6 & 0.021s & 3.041s \\
		\hline
		No.7 & 0.020s & 3.053s \\
		\hline
		No.8 & 0.021s & 0.291s \\
		\hline
		No.9 & 0.020s & N/A \\
		\hline
		No.10 & 0.020s & 0.214s \\
		
		\hline
		\hline

	\end{tabular}
	\smallskip
	
\end{table}

\section{Related Works} \label{sec:Related Works}
There has been a lot of work on applying abduction within logic programming as well as ASP \cite{default_assumption_logic_program,TMS,BARAL,kakas}. Our work shows how abduction can be fruitfully employed within the context of goal-directed ASP to solve problems related to compliance in medicine. 

In other works, Chesani et al. \shortcite{Careflow} describe an approach to perform the conformance verification 
of careflow process executions. The approach translates clinician's workflow 
into a formal language based on computational logic and abductive logic 
programming. A graphical language was developed to specify careflow models. 
Given a set of events that have taken place, expectations are generated which can be 
compared with the actual participants' behavior. If a participant does not 
behave as expected with respect to the model, a violation will be raised by the 
procedure. 

Temporal logic is used to identify ways a treatment can be non-compliant with clinical guidelines \cite{groot_model_checking}. One of possible reasons for a non-compliant treatment is relevant findings are missing in patient data. Those missing relevant parameters in records can be found using the proposed critiquing method.

PROforma \cite{PROforma} is a knowledge representation language that is based 
on classical predicate calculus augmented by first-order logic. In PROforma, an 
application such as a protocol or clinical guideline is modeled as a plan made 
up of one or more tasks. PROforma is able to recognize clinical problems and 
identify possible solutions to them. It can also assess the strengths and 
weaknesses of alternative solutions, yielding a preference order on the set of 
alternatives.

Spiotta et al. \shortcite{Spiotta2015AnswerSP} proposes an approach for analyzing conformance of ASP execution traces for patient treatment with respect to clinical guidelines and basic medical knowledge. Basic medical knowledge may be used to justify deviations from the application of guidelines. 

The above approaches, including ours, capture clinical expertise 
in a form which can be directly applied by one agent (such as a machine) on 
behalf of another (such as an expert physician). 
However, our system does not 
require complete information when performing abductive reasoning, thanks to 
use of negation as failure and ASP. The ability to perform reasoning with
incomplete information is highly valuable in a real world setting.

\section{Conclusions and Future Work} \label{sec:conclusion and future work}
In this paper we show how our physician advisory system for heart failure 
management, extended with abductive reasoning,
can be used to improve adherence to clinical practice guidelines.
With the multitude and complexity of clinical guidelines, the adherence to the guidelines in any one particular 
chronic disease state is an error-prone task even for the best-intentioned 
physicians. Moreover, the fact that complex clinical practice guidelines are 
updated every few years compounds the compliance problem in the heart failure 
management community. Our system takes advantage of the abductive reasoning 
support in s(ASP) and provides physicians with cognitive assistance for the 
rationale behind their treatment choices. Given the patient's data, the 
intended treatment and the abducibles, the physician advisory system is able to 
give the conditions that must hold in order for the intended treatment to be 
legitimate. We showed how our physician advisory system for heart failure 
management uses abductive reasoning to discover the conditions and symptoms 
that a patient must exhibit for a given treatment recommendation to apply. Such 
ability is valuable in helping physicians validate their decisions for 
treatment and provide optimal, evidence-based care in both training sessions and real-world settings. 

Future work is 
planned in the following directions: 

\begin{itemize}
	\item Extending the system to cover 
	the clinical practice guidelines of other co-morbid diseases: 
	HF is usually accompanied by other diseases such as MI, diabetes, 
	etc. 
	
	\item Introducing pruning techniques in the s(ASP) system: 
	the non-ground abducibles result in a massive number of ways to prove 
	the same goal. Addition of suitable pruning operators to s(ASP) 
	will mitigate this problem.
	
	\item Field Trial: conducting a field 
	trial for a large number of patients to validate our system.
	
	\item Integrating with electronic medical records (EMR) and telemedicine platform: an adapter may be needed to enable the physician advisory system to recognize vocabularies of other systems which use different medical terms than 2013 ACCF/AHA Guideline for the Management of Heart Failure. 
	
	\item Displaying the justifications for a query: Researching the best way to show the justifications, including cyclical reasoning due to coinduction.
	
\end{itemize}

\noindent{\bf Acknowledgments:} This research is supported by NSF (Grant No. 
1423419) and the Texas Medical Research Collaborative.

\bibliographystyle{acmtrans}
\bibliography{iclp_2017_arXiv}

\begin{thebibliography}{}

\bibitem[\protect\citeauthoryear{Baral}{Baral}{2003}]{BARAL}
{\sc Baral, C.} 2003.
\newblock {\em Knowledge Representation, Reasoning and Declarative Problem
  Solving}.
\newblock Cambridge University Press.

\bibitem[\protect\citeauthoryear{Cabana, Rand, Powe, and et~al}{Cabana
  et~al\mbox{.}}{1999}]{Why_don't_physicians_follow}
{\sc Cabana, M.~D.}, {\sc Rand, C.~S.}, {\sc Powe, N.~R.}, {\sc and} {\sc
  et~al}. 1999.
\newblock Why don't physicians follow clinical practice guidelines?: A
  framework for improvement.
\newblock {\em JAMA\/}~{\em 282,\/}~15, 1458--1465.
\newblock /data/Journals/JAMA/4708/JRV90041.pdf.

\bibitem[\protect\citeauthoryear{Chen, Marple, Salazar, Gupta, and Tamil}{Chen
  et~al\mbox{.}}{2016}]{iclp2016}
{\sc Chen, Z.}, {\sc Marple, K.}, {\sc Salazar, E.}, {\sc Gupta, G.}, {\sc and}
  {\sc Tamil, L.} 2016.
\newblock A physician advisory system for chronic heart failure management
  based on knowledge patterns.
\newblock {\em {TPLP}\/}~{\em 16,\/}~5-6, 604--618.

\bibitem[\protect\citeauthoryear{Chesani, Mello, Montali, and Storari}{Chesani
  et~al\mbox{.}}{2007}]{Careflow}
{\sc Chesani, F.}, {\sc Mello, P.}, {\sc Montali, M.}, {\sc and} {\sc Storari,
  S.} 2007.
\newblock Testing careflow process execution conformance by translating a
  graphical language to computational logic.
\newblock In {\em Artificial Intelligence in Medicine, 11th Conference on
  Artificial Intelligence in Medicine, {AIME} 2007, Amsterdam, The Netherlands,
  July 7-11, 2007, Proceedings}. 479--488.

\bibitem[\protect\citeauthoryear{Douven}{Douven}{2011}]{abduction}
{\sc Douven, I.} 2011.
\newblock Abduction.
\newblock In {\em The Stanford Encyclopedia of Philosophy\/}, Spring 2011 ed.,
  {E.~N. Zalta}, Ed.

\bibitem[\protect\citeauthoryear{Fox, Johns, and Rahmanzadeh}{Fox
  et~al\mbox{.}}{1998}]{PROforma}
{\sc Fox, J.}, {\sc Johns, N.}, {\sc and} {\sc Rahmanzadeh, A.} 1998.
\newblock Disseminating medical knowledge: the proforma approach.
\newblock {\em Artificial Intelligence in Medicine\/}~{\em 14,\/}~1-2,
  157--182.

\bibitem[\protect\citeauthoryear{Gebser, Kaminski, Kaufmann, and Schaub}{Gebser
  et~al\mbox{.}}{2014}]{Clingo}
{\sc Gebser, M.}, {\sc Kaminski, R.}, {\sc Kaufmann, B.}, {\sc and} {\sc
  Schaub, T.} 2014.
\newblock Clingo = {ASP} + control: Preliminary report.
\newblock {\em CoRR\/}~{\em abs/1405.3694}.

\bibitem[\protect\citeauthoryear{Gelfond and Lifschitz}{Gelfond and
  Lifschitz}{1988}]{The_Stable_Model_Semantics}
{\sc Gelfond, M.} {\sc and} {\sc Lifschitz, V.} 1988.
\newblock The stable model semantics for logic programming.
\newblock MIT Press, 1070--1080.

\bibitem[\protect\citeauthoryear{Go, Mozaffarian, Benjamin, and et~al}{Go
  et~al\mbox{.}}{2013}]{Heart_disease_and_stroke_statistics}
{\sc Go, A.~S.}, {\sc Mozaffarian, D.}, {\sc Benjamin, E.~J.}, {\sc and} {\sc
  et~al}. 2013.
\newblock Heart disease and stroke statistics 2013 update: a report from the
  american heart association.
\newblock Tech. rep.

\bibitem[\protect\citeauthoryear{Groot, Hommersom, Lucas, Merk, ten Teije, van
  Harmelen, and Serban}{Groot et~al\mbox{.}}{2009}]{groot_model_checking}
{\sc Groot, P.}, {\sc Hommersom, A.}, {\sc Lucas, P.~J.}, {\sc Merk, R.-J.},
  {\sc ten Teije, A.}, {\sc van Harmelen, F.}, {\sc and} {\sc Serban, R.} 2009.
\newblock Using model checking for critiquing based on clinical guidelines.
\newblock {\em Artificial Intelligence in Medicine\/}~{\em 46,\/}~1, 19--36.

\bibitem[\protect\citeauthoryear{Group}{Group}{2006}]{Enhancing_the_use_of_clinical_guidelines}
{\sc Group, T. M.~N.} 2006.
\newblock Enhancing the use of clinical guidelines: a social norms perspective.
\newblock {\em Journal of the American College of Surgeons\/}~{\em 202,\/}~5,
  826--836.

\bibitem[\protect\citeauthoryear{Gupta, Bansal, Min, Simon, and Mallya}{Gupta
  et~al\mbox{.}}{2007}]{Coinductive_Logic_Programming}
{\sc Gupta, G.}, {\sc Bansal, A.}, {\sc Min, R.}, {\sc Simon, L.}, {\sc and}
  {\sc Mallya, A.} 2007.
\newblock Coinductive logic programming and its applications.
\newblock In {\em Logic Programming, 23rd International Conference, {ICLP}
  2007, Porto, Portugal, September 8-13, 2007, Proceedings}. 27--44.

\bibitem[\protect\citeauthoryear{Harman}{Harman}{1965}]{harman_abduction}
{\sc Harman, G.~H.} 1965.
\newblock The inference to the best explanation.
\newblock {\em The Philosophical Review\/}~{\em 74,\/}~1, 88--95.

\bibitem[\protect\citeauthoryear{Inoue}{Inoue}{1991}]{default_assumption_logic_program}
{\sc Inoue, K.} 1991.
\newblock Extended logic programs with default assumptions.
\newblock In {\em Logic Programming, Proceedings of the Eigth International
  Conference, Paris, France, June 24-28, 1991}. 490--504.

\bibitem[\protect\citeauthoryear{Jacobs, Kushner, and et~al}{Jacobs
  et~al\mbox{.}}{2013}]{Clinical_Practice_Guideline_Methodology_Summit_Report}
{\sc Jacobs, A.~K.}, {\sc Kushner, F.~G.}, {\sc and} {\sc et~al}. 2013.
\newblock {ACCF/AHA} clinical practice guideline methodology summit report: A
  report of the american college of cardiology foundation/american heart
  association task force on practice guidelines.
\newblock {\em Journal of the American College of Cardiology\/}~{\em 61,\/}~2,
  213--265.
\newblock /data/Journals/JAC/926164/09025.pdf.

\bibitem[\protect\citeauthoryear{Kakas, Kowalski, and Toni}{Kakas
  et~al\mbox{.}}{1992}]{kakas}
{\sc Kakas, A.~C.}, {\sc Kowalski, R.~A.}, {\sc and} {\sc Toni, F.} 1992.
\newblock Abductive logic programming.
\newblock {\em J. Log. Comput.\/}~{\em 2,\/}~6, 719--770.

\bibitem[\protect\citeauthoryear{Marek and Truszczy{\'{n}}ski}{Marek and
  Truszczy{\'{n}}ski}{1999}]{Truszczynski}
{\sc Marek, V.~W.} {\sc and} {\sc Truszczy{\'{n}}ski, M.} 1999.
\newblock Stable models and an alternative logic programming paradigm.
\newblock In {\em The Logic Programming Paradigm: A 25-Year Perspective},
  {K.~R. Apt}, {V.~W. Marek}, {M.~Truszczynski}, {and} {D.~S. Warren}, Eds.
  Springer Berlin Heidelberg, Berlin, Heidelberg, 375--398.

\bibitem[\protect\citeauthoryear{Marple, Bansal, Min, and Gupta}{Marple
  et~al\mbox{.}}{2012}]{Goal-directed_execution_of_answer_set_programs}
{\sc Marple, K.}, {\sc Bansal, A.}, {\sc Min, R.}, {\sc and} {\sc Gupta, G.}
  2012.
\newblock Goal-directed execution of answer set programs.
\newblock In {\em Principles and Practice of Declarative Programming, PPDP'12,
  Leuven, Belgium - September 19 - 21, 2012}. 35--44.

\bibitem[\protect\citeauthoryear{Marple and Gupta}{Marple and
  Gupta}{2012}]{Galliwasp}
{\sc Marple, K.} {\sc and} {\sc Gupta, G.} 2012.
\newblock Galliwasp: {A} goal-directed answer set solver.
\newblock In {\em Logic-Based Program Synthesis and Transformation, 22nd
  International Symposium, {LOPSTR} 2012, Leuven, Belgium, September 18-20,
  2012, Revised Selected Papers}. 122--136.

\bibitem[\protect\citeauthoryear{Marple, Salazar, and Gupta}{Marple
  et~al\mbox{.}}{2016a}]{sasp_paper}
{\sc Marple, K.}, {\sc Salazar, E.}, {\sc and} {\sc Gupta, G.} 2016a.
\newblock {Computing} {Stable} {Models} of {Normal} {Logic} {Programs}
  {Without} {Grounding}.
\newblock Forthcoming.

\bibitem[\protect\citeauthoryear{Marple, Salazar, and Gupta}{Marple
  et~al\mbox{.}}{2016b}]{sasp}
{\sc Marple, K.}, {\sc Salazar, E.}, {\sc and} {\sc Gupta, G.} 2016b.
\newblock s({ASP}). https://sourceforge.net/projects/sasp-system/.

\bibitem[\protect\citeauthoryear{Niemel{\"{a}}}{Niemel{\"{a}}}{1999}]{Ilkka}
{\sc Niemel{\"{a}}, I.} 1999.
\newblock Logic programs with stable model semantics as a constraint
  programming paradigm.
\newblock {\em Ann. Math. Artif. Intell.\/}~{\em 25,\/}~3-4, 241--273.

\bibitem[\protect\citeauthoryear{Satoh and Iwayama}{Satoh and
  Iwayama}{1991}]{TMS}
{\sc Satoh, K.} {\sc and} {\sc Iwayama, N.} 1991.
\newblock Computing abduction by using the {TMS}.
\newblock In {\em Logic Programming, Proceedings of the Eigth International
  Conference, Paris, France, June 24-28, 1991}. 505--518.

\bibitem[\protect\citeauthoryear{Spiotta, Terenziani, and Dupr{\'e}}{Spiotta
  et~al\mbox{.}}{2015}]{Spiotta2015AnswerSP}
{\sc Spiotta, M.}, {\sc Terenziani, P.}, {\sc and} {\sc Dupr{\'e}, D.~T.} 2015.
\newblock Answer set programming for temporal conformance analysis of clinical
  guidelines execution.
\newblock In {\em KR4HC/ProHealth}.

\bibitem[\protect\citeauthoryear{Yancy, Jessup, and et~al}{Yancy
  et~al\mbox{.}}{2013}]{guideline}
{\sc Yancy, C.~W.}, {\sc Jessup, M.}, {\sc and} {\sc et~al}. 2013.
\newblock 2013 {ACCF/AHA} guideline for the management of heart failure: A
  report of the american college of cardiology foundation/american heart
  association task force on practice guidelines.
\newblock {\em Journal of the American College of Cardiology\/}~{\em 62,\/}~16,
  e147.

\end{thebibliography}

\label{lastpage}
\end{document}